\documentclass[11pt,a4paper]{article}
\usepackage[hyperref]{acl2020}
\usepackage{times}
 \usepackage{xcolor}
\usepackage{latexsym}
\usepackage{amssymb}
\usepackage{amsmath}
\usepackage{array}

\usepackage{booktabs} 
\usepackage{graphicx}
\usepackage{setspace}
\usepackage{multirow}
\usepackage{subfigure}
\usepackage{url}
\usepackage{xcolor}
\usepackage{colortbl,booktabs}
\usepackage{microtype}
\aclfinalcopy 
\def\aclpaperid{274}

\newcommand{\modelname}{\textsc{MarCo}}

\title{Multi-Domain Dialogue Acts and Response Co-Generation}

\author{Kai Wang\textsuperscript{1}, Junfeng Tian\textsuperscript{2}, Rui Wang\textsuperscript{2}, Xiaojun Quan\textsuperscript{1}\thanks{\; Corresponding author.}, Jianxing Yu\textsuperscript{1}\\
\textsuperscript{1}School of Data and Computer Science, Sun Yat-sen University, China \\
\textsuperscript{2}Alibaba Group, China\\
\tt wangk73@mail2.sysu.edu.cn, \tt \{quanxj3, yujx26\}@mail.sysu.edu.cn \\
\{\tt tjf141457, masi.wr\}@alibaba-inc.com}

\date{}

\begin{document}
\maketitle

\begin{abstract}
Generating fluent and informative responses is of critical importance for task-oriented dialogue systems.~Existing pipeline approaches generally predict multiple dialogue acts first and use them to assist response generation. There are at least two shortcomings with such approaches.~First, the inherent structures of multi-domain dialogue acts are neglected. Second, the semantic associations between acts and responses are not taken into account for response generation.~To address these issues, we propose a neural co-generation model that generates dialogue acts and responses concurrently.~Unlike those pipeline approaches, our act generation module preserves the semantic structures of multi-domain dialogue acts and our response generation module dynamically attends to different acts as needed.~We train the two modules jointly using an uncertainty loss to adjust their task weights adaptively.~Extensive experiments are conducted on the large-scale MultiWOZ dataset and the results show that our model achieves very favorable improvement over several state-of-the-art models in both automatic and human evaluations.
\end{abstract}

\section{Introduction}
Task-oriented dialogue systems aim to facilitate people with such services as hotel reservation and ticket booking through natural language conversations. Recent years have seen a rapid proliferation of interests in this task from both academia and industry \cite{bordesBW17,budzianowski2018multiwoz,wuMHXSF19}. A standard architecture of these systems generally decomposes this task into several subtasks, including natural language understanding \cite{gupta-etal-2018-semantic-parsing}, dialogue state tracking \cite{zhong2018global} and natural language generation \cite{su-etal-2018-natural}. They can be modeled separately and combined into a pipeline system.

\begin{figure}[t]
  \centering
  \includegraphics[width=7cm]{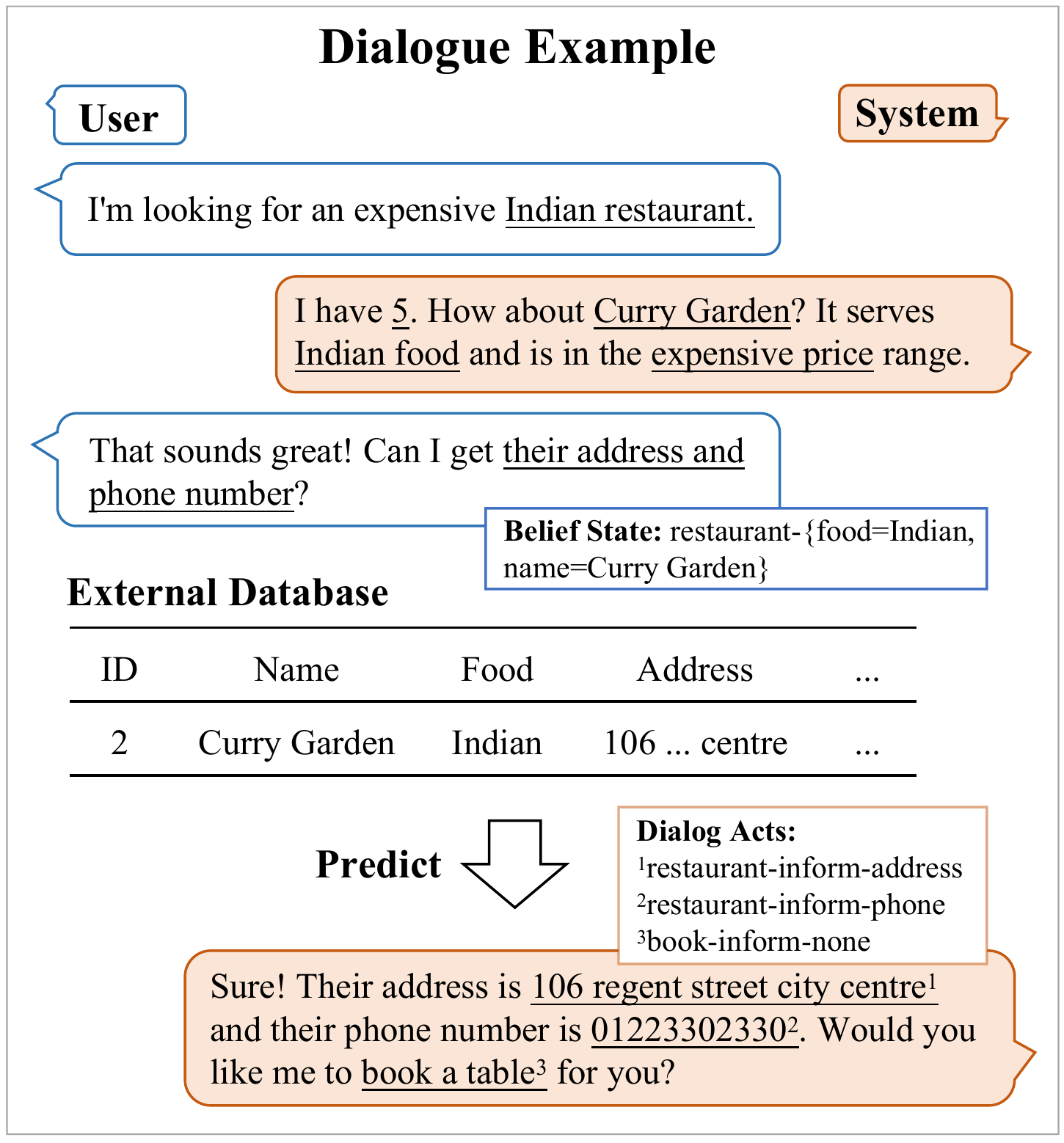}
  \caption{An example of dialogue from the MultiWOZ dataset, where the dialogue system needs to generate a natural language response according to current belief state and related database records.}
  \label{fig:example}
\end{figure}

Figure \ref{fig:example} shows a dialogue example, from which we can notice that the natural language generation subtask can be further divided into dialogue act prediction and response generation \cite{chen-etal-2019-semantically,zhao2019rethinking,wen2017latent}. While the former is intended to predict the next action(s) based on current conversational state and database information, response generation is used to produce a natural language response based on the action(s). 

In order for dialogues to be natural and effective, responses should be fluent, informative, and relevant. Nevertheless, current sequence-to-sequence models often generate uninformative responses like ``I don't know'' \cite{liMRJGG16}, hindering the dialogues to continue or even leading to a failure. Some researchers \cite{pei2019modular,mehri2019structured} sought to combine multiple decoders into a stronger one to avoid such responses, while others \cite{chen-etal-2019-semantically,wen-etal-2015-semantically,zhao2019rethinking,wen2017latent} represent dialogue acts in a global, static vector to assist response generation.

 As pointed out by \citet{chen-etal-2019-semantically}, dialogue acts can be naturally organized in hierarchical structures, which has yet to be explored seriously. Take two acts \textit{station-request-stars} and \textit{restaurant-inform-address} as an example. While the first act rarely appears in real-world dialogues, the second is more often. Moreover, there can be multiple dialogue acts mentioned in a single dialogue turn, which requires the model to attend to different acts for different sub-sequences. Thus, a global vector is unable to capture the inter-relationships among acts, nor is it flexible for response generation especially when more than one act is mentioned.


\begin{figure}
\centering
\includegraphics[width=7cm]{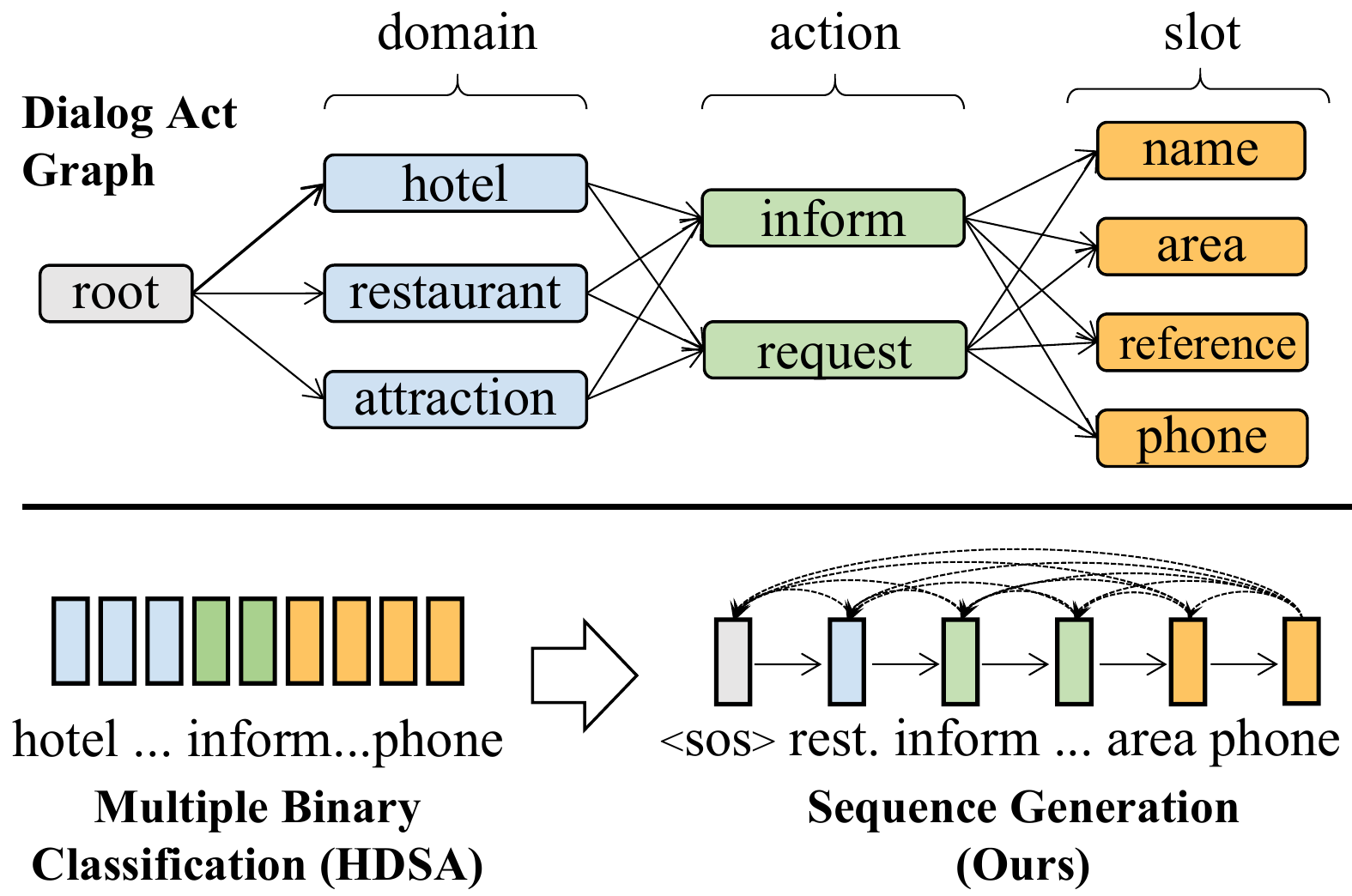}
\caption{Demonstration of hierarchical dialogue act structures (top) and different approaches (bottom) for dialogue act prediction. Classification approaches separately predict each act item (domain, action and slot), while generation approaches treat each act as a token that can be generated sequentially.}
\label{fig:act graph}
\end{figure}

To overcome the above issues, we treat dialogue act prediction as another sequence generation problem like response generation and propose a co-generation model to generate them concurrently. Unlike those classification approaches, act sequence generation not only preserves the inter-relationships among dialogue acts but also allows close interactions with response generation. By attending to different acts, the response generation module can dynamically capture salient acts and produce higher-quality responses. Figure \ref{fig:act graph} demonstrates the difference between the classification and the generation approaches for act prediction.

As for training, most joint learning models rely on hand-crafted or tunable weights on development sets \cite{LiuL17,mrksic-etal-2017-neural,rastogi2018multi}. The challenge here is to combine two sequence generators with varied vocabularies and sequence lengths. The model is sensitive during training and nontrivial to generate an optimal weight. To address this issue, we opt for an uncertainty loss \cite{kendall2018multi} to adaptively adjust the weight according to task-specific uncertainty. We conduct extensive studies on a large-scale task-oriented dataset to evaluate the model. The experimental results confirm the effectiveness of our model with very favorable performance over several state-of-the-art methods.

The contributions of this work include:
\begin{itemize}
  \item We model dialogue act prediction as a sequence generation problem that allows to exploit act structures for the prediction.
  \item We propose a co-generation model to generate act and response sequences jointly, with an uncertainty loss used for adaptive weighting.
  \item Experiments on MultiWOZ verify that our model outperforms several state-of-the-art methods in automatic and human evaluations.
\end{itemize}

\section{Related Work}
Dialogue act prediction and response generation are closely related in general in the research of dialogue systems \cite{chen-etal-2019-semantically,zhao2019rethinking,wen2017latent}, where dialogue act prediction is first conducted and used for response generation. Each dialogue act can be treated as a triple (domain-action-slot) and all acts together are represented in a one-hot vector \cite{wen-etal-2015-semantically,budzianowski2018multiwoz}. Such sparse representation makes the act space very large. To overcome this issue, \newcite{chen-etal-2019-semantically} took into account act structures and proposed to represent the dialogue acts with level-specific one-hot vectors. Each dimension of the vectors is predicted by a binary classifier.

To improve response generation, \newcite{pei2019modular} proposed to learn different \textit{expert} decoders for different domains and acts, and combined them with a \textit{chair} decoder. \newcite{mehri2019structured} applied a cold-fusion method \cite{SriramJSC18} to combine their response decoder with a language model. \newcite{zhao2019rethinking} treated dialogue acts as latent variables and used reinforcement learning to optimize them.~Reinforcement learning was also applied to find optimal dialogue policies in task-oriented dialogue systems \cite{su2017sample,williams2017hybrid} or obtain higher dialog-level rewards in chatting \cite{li2016deep,serban2017deep}. Besides, \newcite{chen-etal-2019-semantically} proposed to predict the acts explicitly with a compact act graph representation and employed hierarchical disentangled self-attention to control response text generation.

Unlike those pipeline architectures, joint learning approaches try to explore the interactions between act prediction and response generation. A large body of research in this direction uses a shared user utterance encoder and train natural language understanding jointly with dialogue state tracking \cite{mrksic-etal-2017-neural,rastogi2018multi}. \newcite{LiuL17} proposed to train a unified network for two subtasks of dialogue state tracking, i.e., knowledge base operation and response candidate selection. \newcite{dialogAct2Vec} showed that joint learning of dialogue act and response benefits representation learning. These works generally demonstrate that joint learning of the subtasks of dialogue systems is able to improve each other and the overall system performance. 

\section{Architecture}

Let ${T} = \{ U_1, R_1, \ldots, U_{t-1}, R_{t-1}, U_t\}$ denote the dialogue history in a multi-turn conversational setting, where $U_i$ and $R_i$ are the $i$-th user utterance and system response, respectively.~${D}=\{d_1,d_2,\ldots,d_n\}$ includes the attributes of related database records for current turn. The objective of a dialogue system is to generate a natural language response $R_t=y_1y_2\ldots y_n$ of $n$ words based on the current belief state and database attributes.

In our framework, dialogue acts and response are co-generated based on the transformer encoder-decoder architecture \cite{vaswani2017attention}.~A standard transformer includes a multi-head attention layer that encodes a value $V$ according to the attention weights from query $Q$ to key $K$, followed by a position-wise feed-forward network ($\mathcal{G}_{f}$):
\begin{equation}
     O = V + \mathcal{G}_{f}(MultiHead(Q,K,V))
\end{equation}
where $Q,K,V,O \in \mathbb{R}^{n\times d}$. In what follows we use $\mathcal{F}(Q, K, V)$ to denote the transformer.

\paragraph{Encoder}
We use $E = \text{Emb}([{T};{D}])$ to represent the concatenated word embeddings of dialogue history $T$ and database attributes $D$. The transformer $\mathcal{F}(Q, K, V)$ is then used to encode $E$ and output its hidden state $H^e$:
\begin{equation}
  H^e = \mathcal{F}(E,E,E)
\end{equation}

\paragraph{Decoder}
At each time step $t$ of response generation, the decoder first computes a self-attention $h_t^{r}$ over already-generated words $y_{1:t-1}$:
\begin{equation}
h_t^{r} =\mathcal{F}(e_{t-1}^{r},e_{1:t-1}^{r},e_{1:t-1}^{r})
\label{attention_1}
\end{equation}
where $e_{t-1}^{r}$ is the embedding of the $(t-1)$-th generated word and $e_{1:t-1}^{r}$ is an embedding matrix of $e_{1}^{r}$ to $e_{t-1}^{r}$. Cross-attention from $h_t^{r}$ to dialogue history $T$ is then executed:
\begin{equation}
c_t^{r} =\mathcal{F}(h_t^{r},H^e,H^e)
\label{attention_2}
\end{equation}
The resulting vectors of Equations \ref{attention_1} and \ref{attention_2}, $h_t^{r}$ and $c_t^{r}$, are concatenated and mapped to a distribution of vocabulary size to predict next word:
\begin{equation}\label{eq:resp prob}
  \begin{split}
    p(y_t|y_{1:t-1}) = \text{softmax}(W_r[c_{t}^r;h_{t}^r])
  \end{split}
  \end{equation}

\section{The \modelname{} Approach}
\begin{figure*}[]
  \centering
  \includegraphics[width=14cm]{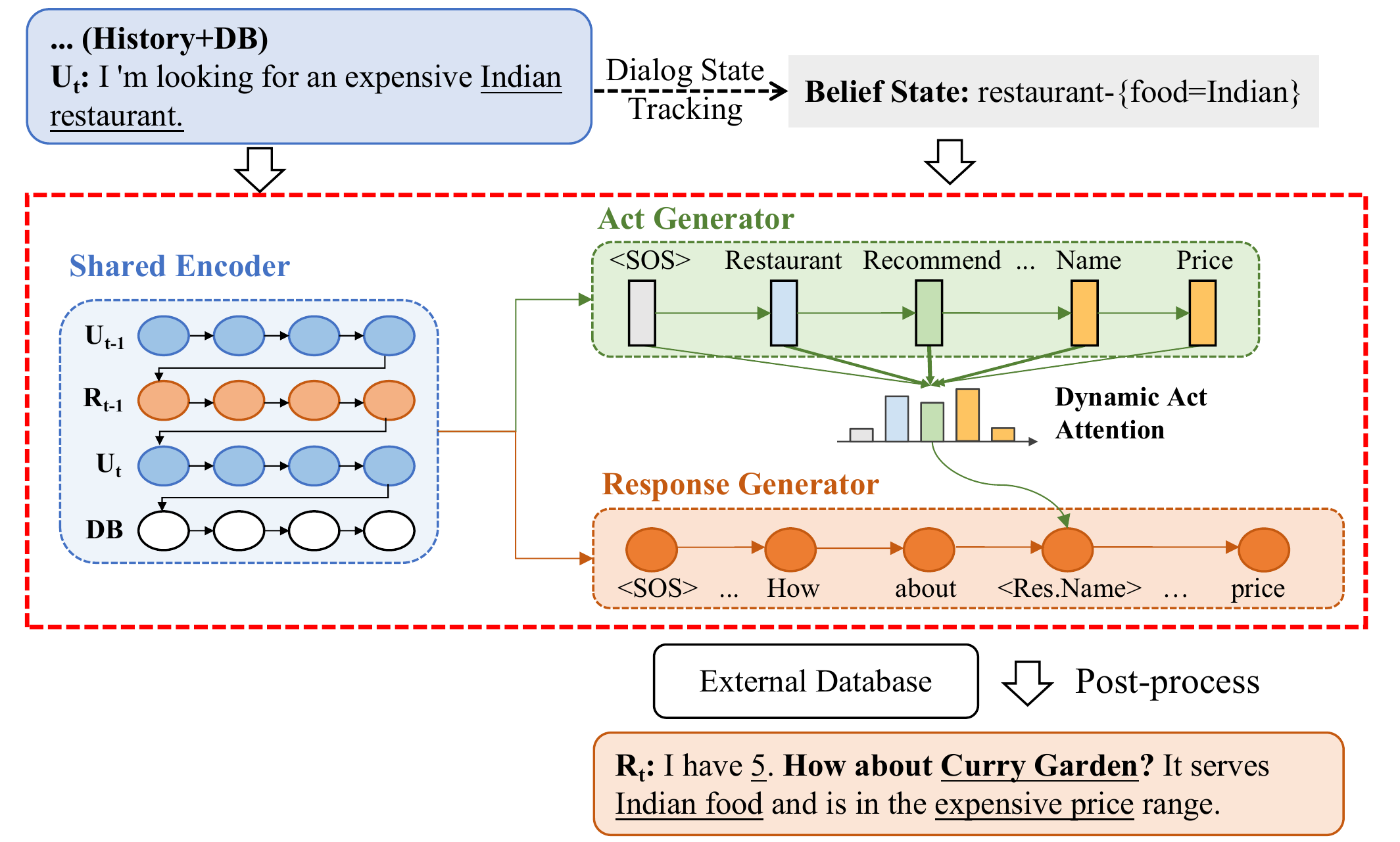}
  \caption{Architecture of the proposed model for act and response co-generation, where act and response generators share the same encoder. The response generator is allowed to attend to different act hidden states as needed using dynamic act attention. The two generators are trained jointly and optimized by the uncertainty loss.}
  \label{fig:DARG}
\end{figure*}

Based on the above encoder-decoder architecture, our model is designed to consist of three components, namely, a shared encoder, a dialogue act generator, and a response generator.~As shown in Figure \ref{fig:DARG}, instead of predicting each act token individually and separately from response generation, our model aims to generate act sequence and response concurrently in a joint model which is optimized by the uncertainty loss \cite{kendall2018multi}.
\vspace{-0.45cm}
\subsection{Dialogue Acts Generation}
Dialogue acts can be viewed as a semantic plan for response generation. As shown in Figure \ref{fig:act graph}, they can be naturally organized in hierarchical structures, including domain level, action level, and slot level. Most existing methods treat dialogue acts as triples represented in one-hot vectors and predict the vector values with binary classifiers \cite{wen-etal-2015-semantically,budzianowski2018multiwoz}. Such representations ignore the inter-relationships and associations among acts, domains, actions and slots.~For example, the slot \textit{area} may appear in more than one domain. Unlike them, we model the prediction of acts as a sequence generation problem, which takes into consideration the structures of acts and generates each act token conditioned on its previously-generated tokens. In this approach, different domains are allowed to share common slots and the search space of dialogue act is greatly reduced.

The act generation starts from a special token ``$\langle \text{SOS} \rangle$'' and produces dialogue acts $A =a_1a_2\ldots a_n$ sequentially. During training, the act sequence is organized by domain, action and slot, while items at each level are arranged in dictionary order, where identical items are merged.~When decoding each act token, we first represent the current belief state with an embedding vector $v_{b}$ and add it to each act word embedding $e_t^{a}$ as:
\begin{equation}
    u_t^{a}  =W_bv_{b}+e_t^{a}.
\end{equation}
Finally, the decoder of Section 3.2 is used to generate hidden states $H^{a}$ and act tokens accordingly.

\subsection{Acts and Response Co-Generation} \label{subsec:mtl}
Dialogue acts and responses are closely related in dialogue systems. On one hand, system responses are generated based on dialogue acts. On the other, their shared information can improve each other through joint learning.

\paragraph{Shared Encoder}~Our dialogue act generator and response generator share one same encoder and input, but having different masking strategies for the input to focus on different information.~In particular, only the current utterance is kept for act generation, while the entire history utterances are used for response generation.\footnote{Empirical evidences show that act generation is more related to the current utterance, while response generation benefits more from long dialogue history.} 

\paragraph{Dynamic Act Attention} A response usually corresponds to more than one dialogue act in multi-domain dialogue systems. Nevertheless, existing methods mostly use a static act vector to represent all the acts, and add the vector to each response token representation. They ignore the fact that different subsequences of a response may need to attend to different acts. To address this issue, we compute dynamic act attention $o_t^r$ from the response to acts when generating a response word:
\begin{equation}\label{eq:act attention}
o_t^r=\mathcal{F}(h_t^{r},H^a,H^a)
\end{equation}
where $h_t^{r}$ is the current hidden state produced by Equation \ref{attention_1}. Then, we combine $o_t^r$ and $h_t^r$ with response-to-history attention $c_t^r$ (by Equation \ref{attention_2}) to estimate the probabilities of next word:
\begin{equation}\label{eq:resp prob}
\begin{aligned}
  p(y_t|y_{1:t-1}) = \text{softmax}(W_r[h_{t}^r;c_{t}^r;o_{t}^r])
\end{aligned}
\end{equation}

\paragraph{Uncertainty Loss}
The cross-entropy function is used to measure the generation losses, $\mathcal{L}_a(\theta)$ and $\mathcal{L}_r(\theta)$, of dialogue acts and responses, respectively:
\begin{align}
  &\mathcal{L}_a(\theta)= -\sum\limits_{j=1}^{T_a}\log p(a^{*(i)}_j|a_{1:j-1}^{(i)},T,D,v_{b}) \\
  &\mathcal{L}_r(\theta) = -\sum\limits_{j=1}^{T_r}\log p(y^{*(i)}_j|y_{1:j-1}^{(i)},T,D,A)
\end{align}
where the ground-truth tokens of acts and response of each turn are represented by $A^*$ and $Y^*$, while the predicted tokens by $A$ and $Y$.

To optimize the above functions jointly, a general approach is to compute a weighted sum like:
\begin{equation}\label{eq:weight sum loss}
  \mathcal{L}(\theta)=\alpha \mathcal{L}_a(\theta)+(1-\alpha)\mathcal{L}_r(\theta)
\end{equation}
However, dialogue acts and responses vary seriously in sequence length and vocabulary size, making the weight $\alpha$ unstable to tune. Instead, we opt for an uncertainty loss \cite{kendall2018multi} to adjust it adaptively:
\begin{equation}
  \mathcal{L}(\theta,\sigma_1,\sigma_2)=\frac{1}{2\sigma_1^2}\mathcal{L}_a(\theta)+\frac{1}{2\sigma_2^2}\mathcal{L}_r(\theta)+\log\sigma_1^2\sigma_2^2 
\end{equation}
where $\sigma_1$ and $\sigma_2$ are two learnable parameters. The advantage of this uncertainty loss is that it models the homoscedastic uncertainty of each task and provides task-dependent weight for multi-task learning \cite{kendall2018multi}. Our experiments also confirm that it leads to more stable weighting than the traditional approach (Section \ref{sec:uncertainty loss}). 

\section{Experiments}
\subsection{Dataset and Metrics}
MultiWOZ 2.0 \cite{budzianowski2018multiwoz} is a large-scale multi-domain conversational datatset consisting of thousands of dialogues in seven domains. 
For fair comparison, we use the same validation set and test set as previous studies \cite{chen-etal-2019-semantically,zhao2019rethinking,budzianowski2018multiwoz}, each set including 1000 dialogues.\footnote{There are only five domains (\emph{restaurant}, \emph{hotel}, \emph{attract}, \emph{taxi}, \emph{train}) of dialogues in the test set as the other two (\emph{hospital}, \emph{police}) have insufficient dialogues.} We use the \emph{Inform Rate} and \emph{Request Success} metrics to evaluate dialog completion, with one measuring whether a system has provided an appropriate entity and the other assessing if it has answered all requested attributes. Besides, we use BLEU \cite{papineni2002bleu} to measure the fluency of generated responses. To measure the overall system performance, we compute a combined score: $(\textit{Inform Rate}+\textit{Request Success})\times0.5+\textit{BLEU}$ as before \cite{budzianowski2018multiwoz,mehri2019structured,pei2019modular}.

\subsection{Implementation Details}
The implementation\footnote{\url{https://github.com/InitialBug/MarCo-Dialog}} is on a single Tesla P100 GPU with a batch size of 512.~The dimension of word embeddings and hidden size are both set to 128.~We use a 3-layer transformer with 4 heads for the multi-head attention layer. For decoding, we use a beam size of 2 to search for optimal results, and apply trigram avoidance \cite{paulus2018a} to fight trigram-level repetition.~During training, we first train the act generator for 10 epochs for warm-up and then optimize the uncertainty loss with the Adam optimizer \cite{KingmaB14}.

\begin{table*}[h]
    \setlength{\belowcaptionskip}{-0.3cm}
	\centering
	\resizebox{2\columnwidth}{!}{%
	\begin{tabular}{lp{3cm}p{2cm}<{\centering}p{2cm}<{\centering}p{2cm}<{\centering}p{2.6cm}<{\centering}}
		\toprule
        Dialog Act&Model     & Inform& Success&BLEU&Combined Score  \\
        \midrule
        \multirow{4}{*}{Without Act}
        &LSTM &71.29&60.96&18.80&84.93\\
        &Transformer &71.10&59.90&19.10&84.60\\
        &TokenMoE &75.30&59.70&16.81&84.31\\
        &Structured Fusion &82.70&72.10&16.34&93.74\\
           \midrule
        \multirow{3}{*}{One-hot Act}
         &SC-LSTM  & 74.50 & 62.50 & 20.50&89.00  \\
          &HDSA (\modelname{}{})  &76.50&62.30	& 21.85 &91.25\\
          &HDSA  &82.90&68.90& \textbf{23.60} & 99.50\\
             \midrule
          \multirow{2}{*}{Sequential Act}
         & \modelname{}{}   & 90.30	& 75.20 &19.45& 102.20\\
         & \modelname{}{} (BERT) & \textbf{92.30}	& \textbf{78.60}	&20.02&\textbf{105.47}\\
        \bottomrule
	\end{tabular}
	}
    \caption{Overall results on the MultiWOZ 2.0 dataset.} 
    \label{tab:overall}
\end{table*} 

\subsection{Baselines}
A few mainstream models are used as baselines for comparison with our neural co-generation model (\textsc{MarCo}), being categorized into three categories:
\vspace{-0.53cm}
\begin{itemize}
\setlength\itemsep{-0.05cm}
    \item
    \textbf{Without Act}.~Models in this category directly generate responses without act prediction, including LSTM \cite{budzianowski2018multiwoz}, Transformer \cite{vaswani2017attention}, TokenMoE \cite{pei2019modular} and Structured Fusion \cite{mehri2019structured}.

    \item
    \textbf{One-Hot Act}.~In SC-LSTM \cite{wen-etal-2015-semantically}, dialogue acts are treated as triples and information flow from acts to response generation is controlled by gates.~HDSA \cite{chen-etal-2019-semantically} is a strong two-stage model, which relies on BERT \cite{devlin2019bert} to predict a one-hot act vector for response generation.
    \item
    \textbf{Sequential Act}.~Since our model does not rely on BERT, to make a fair comparison with HDSA, we design the experiments from two aspects to ensure they have the same dialogue act inputs for response generation. First, the act sequences produced by our co-generation model are converted into one-hot vectors and fed to HDSA. Second, the predicted one-hot act vectors by BERT are transformed into act sequences and passed to our model as inputs.
\end{itemize}

\subsection{Overall Results}
The overall results are shown in Table \ref{tab:overall}, in which HDSA (\modelname{}{}) means HDSA using \modelname{}{}'s dialogue act information, and \modelname{}{} (BERT) means \modelname{}{} based on BERT's act prediction. From the table we can notice that our co-generation model (\modelname{}{}) outperforms all the baselines in \emph{Inform Rate}, \emph{Request Success}, and especially in \emph{combined score} which is an overall metric. By comparing the two HDSA models, we can find HDSA derives its main performance from the external BERT, which can also be used to improve our \modelname{}{} considerably (\modelname{}{} (BERT)). These results confirm the success of \modelname{}{} by modeling act prediction as a generation problem and training it jointly with response generation. 

Another observation is that despite its strong overall performance, \modelname{}{} shows inferior BLEU performance to the two HDSA models. The reason behind this is studied and analyzed in human evaluation (Section \ref{sec:human eval}), showing that our model often generates responses inconsistent with references but favored by human judges.

\begin{figure}[t]
    \centering
    \includegraphics[width=1\columnwidth]{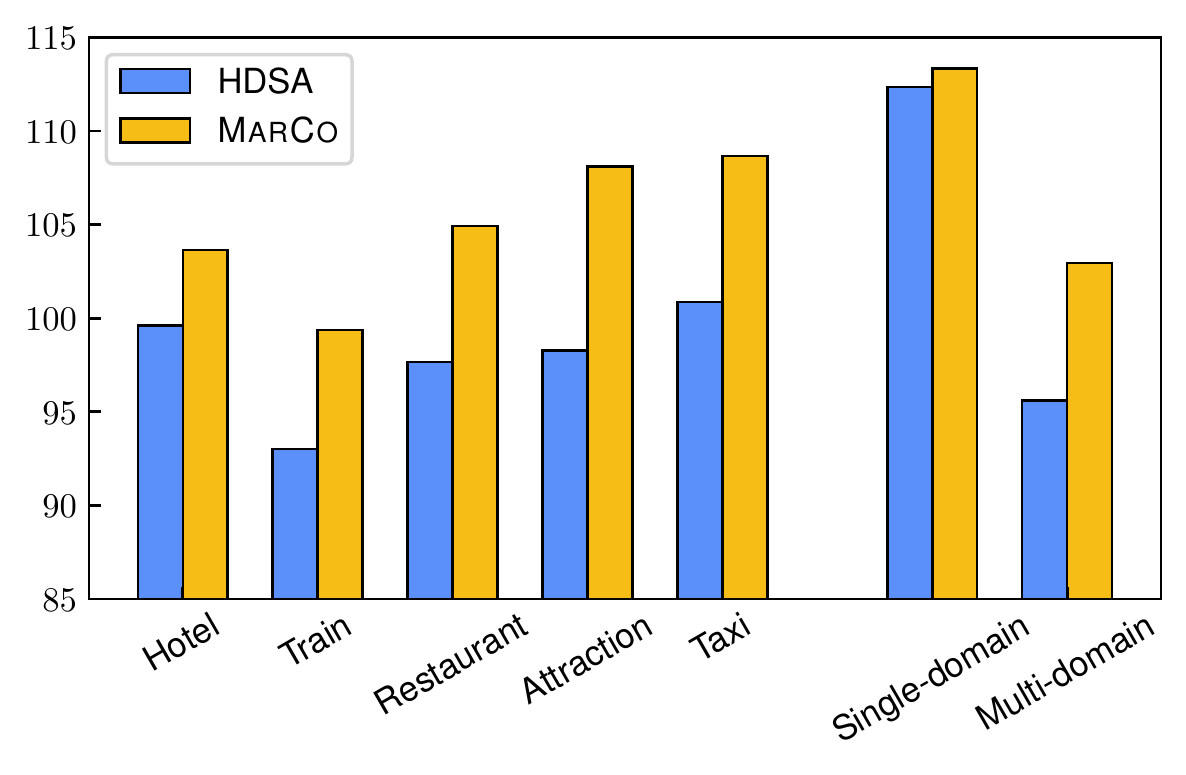}
    \caption{Combined score of \modelname{}{} vs.~HDSA across different domains. If a dialogue involves more than one domain, it is copied into each domain. Single-domain includes dialogues with only one domain mentioned, while the rest belongs to the multi-domain.}
    \label{fig:domain}
\end{figure}

The performance of our model across different domains is also compared against HDSA. The average number of turns is 8.93 for single-domain dialogues and 15.39 for multi-domain dialogues \cite{budzianowski2018multiwoz}. As in Figure \ref{fig:domain}, our model shows superior performance to HDSA across all domains. The results suggest that \modelname{}{} is good at dealing with long dialogues.


\paragraph{Results on MultiWOZ 2.1} We also conducted experiments on MultiWOZ 2.1 \cite{eric2019multiwoz}, which is an updated version of MultiWOZ 2.0. As shown in Table \ref{tab:multiwoz21}, the overall results are consistent with that on MultiWOZ 2.0.

\begin{table}[t]
    \centering
    \setlength{\tabcolsep}{0.2em}
    \begin{tabular}{lcccc}
		\toprule
        Model     & Inform & Success & BLEU & Score \\
        \midrule
        Transformer & 72.50	& 52.70	& 19.08 & 81.68 \\
        HDSA        & 86.30 & 70.60 & \textbf{22.36} & 100.81 \\
        \modelname{}{}  &  91.50 & 76.10	& 18.52	& 102.32 \\
        \modelname{}{} (BERT) & \textbf{92.50}	& \textbf{77.80} & 19.54 & \textbf{104.69}\\
        \bottomrule
	\end{tabular}
    \caption{Overall results on the MultiWOZ 2.1 dataset.}
    \label{tab:multiwoz21}
\end{table}

\section{Further Analysis}
More thorough studies and analysis are conducted in this section, trying to answer three questions: (1) How is the performance of our act generator in comparison with existing classification methods? (2) Can our joint model successfully build semantic associations between acts and responses? (3) How does the uncertainty loss contribute to our co-generation model?

\subsection{Dialogue Act Prediction}
To evaluate the performance of our act generator, we compare it with several baseline methods mentioned in \cite{chen-etal-2019-semantically}, including BiLSTM, Word-CNN, and 3-layer Transformer. We use \modelname{}{} to represent our act generator which is trained jointly with the response generator, and use Transformer (GEN) to denote our act generator without joint training. From Table \ref{tab:act prediction performance}, we notice that the separate generator, Transformer (GEN), performs much better than BiLSTM and Word-CNN, but comparable with Transformer. But after trained jointly with the response generator, \modelname{}{} manages to show the best performance, confirming the effect of the co-generation.

\begin{table}[]
	\centering
	\begin{tabular}{p{6cm}p{0.7cm}<{\centering}}
		\toprule
         Method     &F1  \\
        \midrule
        BiLSTM    &71.4\\
        Word-CNN  &71.5\\
        Transformer  &73.1\\
        \rowcolor{gray!10} Transformer (GEN)&73.2\\
        \rowcolor{gray!10} \modelname{}{} & \textbf{73.9} \\
        \bottomrule
	\end{tabular}
    \caption{Results of different act generation methods, where BiLSTM, Word-CNN and Transformer are baselines from \cite{chen-etal-2019-semantically}. \modelname{}{} is our act generator trained jointly with the response generator and Transformer (GEN) is that without joint training.}
    \label{tab:act prediction performance}
\end{table}

\subsection{Joint vs. Pipeline}

\begin{table}[t]
    \setlength{\belowcaptionskip}{-0.3cm}
	\centering
	\begin{tabular}{lp{0.9cm}<{\centering}p{0.75cm}<{\centering}p{0.75cm}<{\centering}p{1.3cm}<{\centering}}
		\toprule
        Model   & Inform & Succ & BLEU &Combined  \\
        \hline
         HDSA&82.9&68.9&23.60&99.50\\
         Pipeline$_1$  &84.3&54.4&16.00&85.35\\
         Pipeline$_2$ &86.6	&66.0	&18.31&94.61\\
        \rowcolor{gray!10} Joint &\textbf{90.3}	& \textbf{75.2}&\textbf{19.45}&\textbf{102.20}\\
        \bottomrule
	\end{tabular}
    \caption{Results of response generation by joint and pipeline models, where Pipeline$_1$ and Pipeline$_2$ represent two pipeline approaches with or without using dynamic act attention. The performance of HDSA, as the best pipeline model, is provided for comparison.}
    \label{tab:joint learning}
\end{table}

\begin{figure}[t]
    \centering
    \includegraphics[width=\columnwidth]{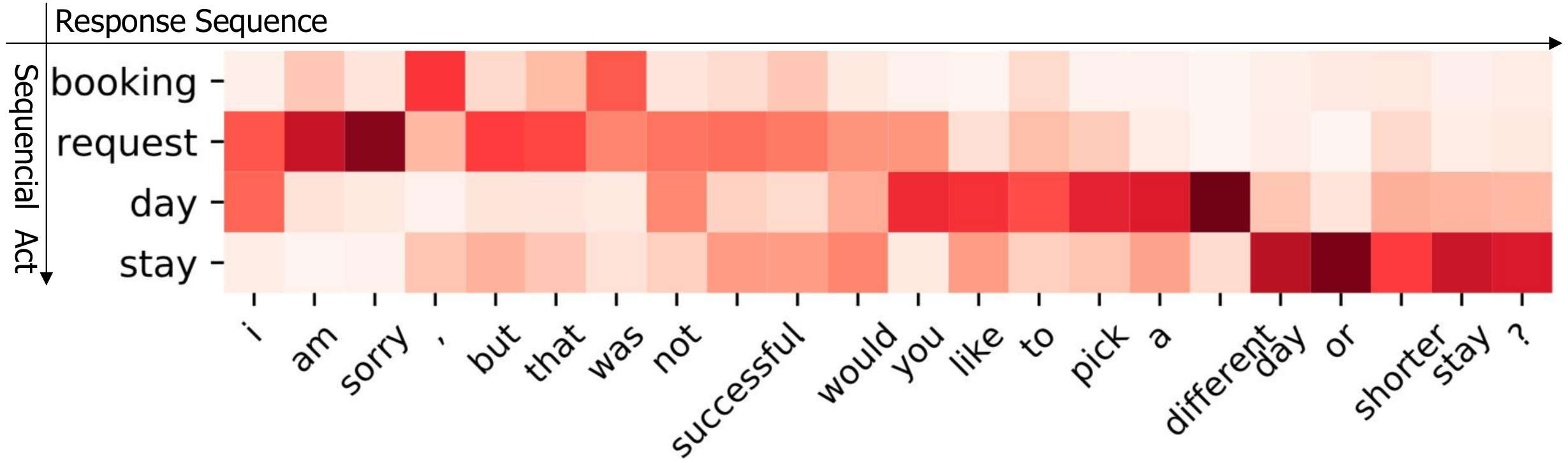}
    \caption{An illustrative example of the dynamic act attention mechanism. Response (row) subsequence can attend to the act (column) token ``day'' or ``stay'' as needed when generating a response asking about picking a different day or shorter stay.}
    \label{tab:heat map}
\end{figure}

To study the influence of the joint training and the dynamic act attention on response generation, we implement two pipeline approaches for comparison.
We first train our act generator separately from response generation. Then, we keep its parameters fixed and train the response generator. The first baseline is created by replacing the dynamic act attention (Equation \ref{eq:act attention}) with an average of the act hidden states, while the second baseline uses the dynamic act attention. As shown in Table \ref{tab:joint learning}, Pipeline$_2$ with dynamic act attention is superior to Pipeline$_1$ without it in all metrics, but inferior to the joint approach. Our joint model also surpasses the currently state-of-the-art pipeline system HDSA, even HDSA uses BERT.
We find that by utilizing sequential acts, the dynamic act attention mechanism helps the response generator capture the local information by attending to different acts.

An illustrative example is shown in Figure \ref{tab:heat map}, where the response generator can attend to the local information such as ``day'' and ``stay'' as needed when generating a response asking about picking a different day or shorter stay.
We reckon that by utilizing sequential acts, response generation benefits in two ways. First, the dynamic act attention allows the generator to attend to different acts when generating a subsequence. Second, the joint training makes the two stages interact with each other, easing error propagation of pipeline systems.

\subsection{Uncertainty Loss}\label{sec:uncertainty loss}

\begin{figure}[t]
    \setlength{\abovecaptionskip}{-0.1cm}
    \centering
    \includegraphics[width=7cm]{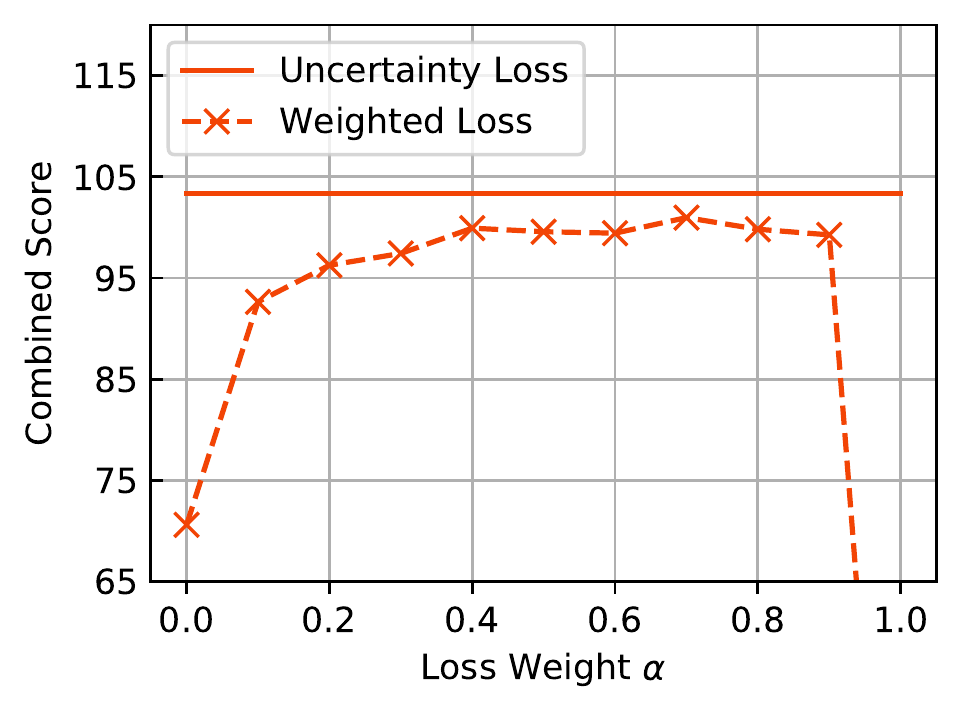}
    \caption{Performance of the uncertainty loss and the weighted-sum loss on the development dataset.}
    \label{tab:uncertainty loss} 
\end{figure}


We opt for an uncertainty loss to optimize our joint model, rather than a traditional weighted-sum loss. To illustrate their difference, we conduct an experiment on the development set. For the traditional loss (Equation \ref{eq:weight sum loss}), we run for each weight from 0 to 1 stepped by 0.1. Note that since the weights, $\sigma_1$ and $\sigma_2$, in the uncertainty loss are not hyperparameters	but learned internally to each batch, we only record the best score within each round without giving the values of $\sigma_1$ and $\sigma_2$. As shown in Figure \ref{tab:uncertainty loss}, the uncertainty loss can learn adaptive weights with consistently superior performance.



\section{Human Evaluation}\label{sec:human eval}

\begin{figure}[t]
    \setlength{\belowcaptionskip}{-0.2cm}
    \centering
    \includegraphics[width=8cm]{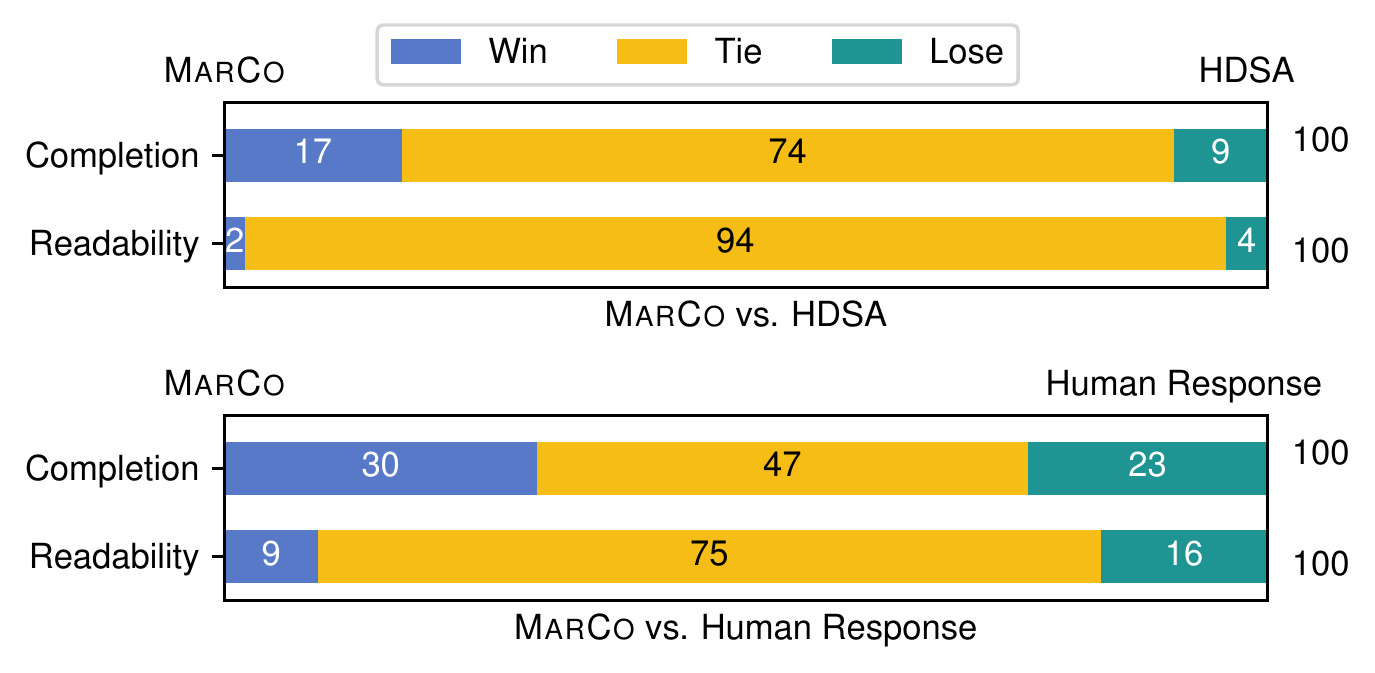}
    \caption{Results of human study in response quality.  Two groups of systems are studied, where the top figure corresponds to results of \modelname{}{} vs. HDSA and the bottom figure represents \modelname{}{} vs. Human Response (ground-truth). ``Win'', ``Tie'' or ``Lose'' respectively indicate the proportions that our \modelname{}{} system wins over, ties with or loses to its counterpart. }
    \label{tab:human evaluation}
\end{figure}

We conduct a human study to evaluate our model by crowd-sourcing.\footnote{The annotation results are available at \url{https://github.com/InitialBug/MarCo-Dialog/tree/master/human\_evaluation}} For this purpose we randomly selected 100 sample dialogues (742 turns in total) from the test dataset and constructed two groups of systems for comparison: \modelname{}{} vs. HDSA and \modelname{}{} vs. Human Response, where Human Response means the reference responses. Responses generated by each group were randomly assigned in pairs to 3 judges, who ranked them according to their completion and readability \cite{chen-etal-2019-semantically,zhang2019dialogpt}. \emph{Completion} measures if the response correctly answers a user query, including relevance and informativeness. \emph{Readability} reflects how fluent, natural and consistent the response is.

\begin{figure}[t]
  \setlength{\belowcaptionskip}{-0.2cm}
    \centering
    \includegraphics[width=\columnwidth]{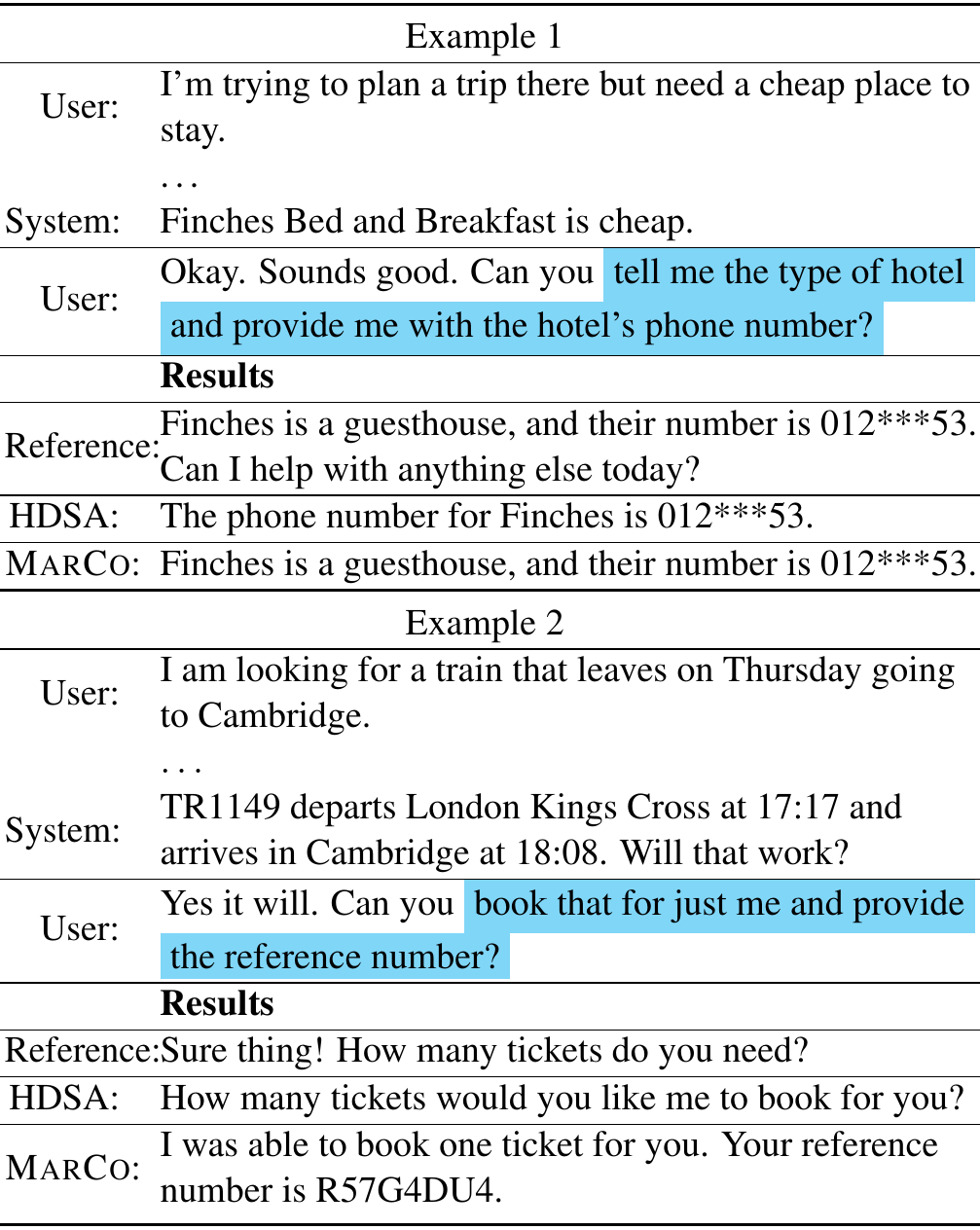}
    \caption{Two examples to show that \modelname{}{} successfully improves the dialogue system by generating relevant and informative responses. }
    \label{tab:case_study_exm}
\end{figure}

The results of this study are shown in Figure \ref{tab:human evaluation}, where ``Win'', ``Tie'' or ``Lose'' mean our \modelname{}{} system wins over, ties with or loses to its counterpart, respectively. From the results we note that \modelname{}{} outperforms HDSA and Human Response in completion, and ties 94\% with HDSA in readability while underperforming Human Response. Overall speaking, \modelname{}{} is superior to HDSA and comparable with Human Response. We further analyzed the bad cases of our model in readability and found that our model slightly suffers from token level repetition, a problem that can be solved by methods like the coverage mechanism \cite{mi2016coverage,tu2016modeling}. In completion, our model can understand the users' need and tends to provides them more relevant information, so that they can finish their goals in shorter turns.

We present two examples in Figure \ref{tab:case_study_exm}. In the first example, the user requests the hotel type while HDSA ignores it. The user requests to book one ticket in the second example, yet both HDSA and Human Response ask about the number once again. In contrast, our model directly answers the questions with correct information. To sum up, \modelname{}{} successfully improves the dialogue system by generating relevant and informative responses.

\section{Conclusion}
In this paper, we presented a novel co-generation model for dialogue act prediction and response generation in task-oriented dialogue systems. Unlike previous approaches, we modeled act prediction as a sequence generation problem to exploit the semantic structures of acts and trained it jointly with response generation via dynamic attention from response generation to act prediction. To train this joint model, we applied an uncertainty loss for adaptive weighting of the two tasks. Extensive studies were conducted on a large-scale task-oriented dataset to evaluate the proposed model, and the results confirm its effectiveness with very favorable performance over several state-of-the-art methods.

\section*{Acknowledgments}

The work was supported by the Fundamental Research Funds for the Central Universities (No.19lgpy220 and No.19lgpy219), the Program for Guangdong Introducing Innovative and Entrepreneurial Teams (No.2017ZT07X355) and the National Natural Science Foundation of China (No.61906217). Part of this work was done when the first author was an intern at Alibaba.

\bibliography{acl2020}
\bibliographystyle{acl_natbib}

\end{document}